\documentclass[10pt,twocolumn,letterpaper]{article}

\usepackage{cvpr}
\usepackage{times}
\usepackage{epsfig}
\usepackage{graphicx}
\usepackage{amsmath}
\usepackage{amssymb}
\usepackage{tabularx}
\newcolumntype{C}[1]{>{\centering\let\newline\\\arraybackslash\hspace{0pt}}p{#1}}

\def\paratitle{} 

\usepackage[pagebackref=true,breaklinks=true,letterpaper=true,colorlinks,bookmarks=false]{hyperref}

\usepackage[numbers,sort]{natbib}
\usepackage{xcolor, colortbl}
\definecolor{Gray}{gray}{0.9}
\newcolumntype{g}{>{\columncolor{Gray}}c}
\newcommand{\suha}[1]{{\color{red} #1}}    
\newcommand{\sh}[1]{{\color{magenta} #1}}    

\cvprfinalcopy 
\newcommand{\I}{\mathcal{I}}
\newcommand{\V}{\mathcal{V}}
\newcommand{\x}{\mathbf{x}}
\newcommand{\y}{\mathbf{y}}
\newcommand{\vvs}{\mathbf{V}}
\newcommand{\vv}{\mathbf{v}}

\newcommand{\R}{\mathbb{R}}
\newcommand{\att}{\boldsymbol{\alpha}}
\newcommand{\W}{W}
\newcommand{\fenc}{f_{\text{enc}}}

\newcommand{\fdec}{f_{\text{dec}}}
\newcommand{\penc}{\theta_{\text{enc}}}
\newcommand{\pdec}{\theta_{\text{dec}}}
\newcommand{\node}{\mathcal{S}}
\newcommand{\edge}{\mathcal{E}}
\newcommand{\spl}{\mathcal{L}}
\newcommand{\spll}{\mathbf{z}}
\newcommand{\s}{\mathbf{s}}

\def\cvprPaperID{3568} 

\ifcvprfinal\fi
\begin{document}

\title{Weakly Supervised Semantic Segmentation using Web-Crawled Videos\vspace{-0.2cm}}

\author{
\begin{tabular}{ccccc}
Seunghoon Hong\textsuperscript{\dag}& Donghun Yeo\textsuperscript{\dag}& Suha Kwak\textsuperscript{\ddag}& Honglak Lee\textsuperscript{\S}& Bohyung Han\textsuperscript{\dag}
\vspace{0.05cm}
\end{tabular}\\
\begin{tabular}{@{}C{4.1cm}@{}C{4.1cm}@{}C{4.1cm}}
\textsuperscript{\dag}POSTECH & \textsuperscript{\ddag}DGIST & \textsuperscript{\S}University of Michigan \\
Pohang, Korea & Daegu, Korea & Ann Arbor, MI, USA\\  
\end{tabular}\\
\begin{tabular}{ccc}
{\tt\small \{maga33, hanulbog, bhhan\}@postech.ac.kr} & {\tt\small skwak@dgist.ac.kr} & {\tt\small honglak@umich.edu} 
\end{tabular}
\vspace{-0.2cm}
}

\maketitle
\thispagestyle{empty}


\begin{abstract}
\vspace*{-0.1in}
We propose a novel algorithm for weakly supervised semantic segmentation based on image-level class labels only.
In weakly supervised setting, it is commonly observed that trained model overly focuses on  discriminative parts rather than the entire object area.
Our goal is to overcome this limitation with no additional human intervention by retrieving videos relevant to target class labels from web repository, and generating segmentation labels from the retrieved videos to simulate strong supervision for semantic segmentation.
During this process, we take advantage of image classification with discriminative localization technique to reject false alarms in retrieved videos and identify relevant spatio-temporal volumes within retrieved videos.
Although the entire procedure does not require any additional supervision, the segmentation annotations obtained from videos are sufficiently strong to learn a model for semantic segmentation.
The proposed algorithm substantially outperforms existing methods based on the same level of supervision and is even as competitive as the approaches relying on extra annotations.
\vspace*{-0.1in}
\end{abstract}

\section{Introduction}
\label{sec:intro}

Semantic segmentation has recently achieved prominent progress thanks to Deep Convolutional Neural Networks (DCNNs)~\cite{Fcn,Deeplabcrf,Crfrnn,Vemulapalli2016,Lin16,Qi2016}. 
The success of DCNNs heavily depends on the availability of a large-scale training dataset, where annotations are given manually in general.
In semantic segmentation, however, annotations are in the form of pixel-wise masks, and collecting such annotations for a large number of images demands tremendous effort and cost.
Consequently, accurate and reliable segmentation annotations are available only for a small number of classes.
Fully supervised DCNNs for semantic segmentation are thus limited to those classes and hard to be extended to many other classes appearing in real world images.

Weakly supervised approaches have been proposed to alleviate this issue by leveraging a vast amount of weakly annotated images.
Among several types of weak supervision for semantic segmentation, image-level class label has been widely used~\cite{Wsl,Wssl,Fcmil,Ccnn,sec} as it is readily available from existing image databases~\cite{Imagenet,Pascalvoc}.
The most popular approach to generating pixel-wise labels from an image-level label is self-supervised learning based on the joint estimation of segmentation annotation and model parameters~\cite{Wsl,Fcmil,Boxsup, scribblesup}. 
However, since there is no way to measure the quality of estimated annotations, these approaches easily converge to suboptimal solutions. 
To remedy this limitation, other types of weak supervision have been employed in addition to image-level labels, \eg, bounding box~\cite{Boxsup,Wssl}, scribble~\cite{scribblesup}, prior meta-information~\cite{Ccnn}, and segmentation ground-truths of other classes~\cite{transfernet}.
However, they often require additional human intervention to obtain extra supervision~\cite{Boxsup,Wssl,transfernet} or employ domain-specific knowledge that may not be well-generalized to other classes~\cite{Ccnn}.

The objective of this work is to overcome the inherent limitation in weakly supervised semantic segmentation without additional human supervision. 
Specifically, we propose to retrieve \textit{videos} from the Web and use them as an additional source of training data,
since temporal dynamics in video offers rich information to distinguish objects from background and estimate their shapes more accurately.
More importantly, our video retrieval process is performed  fully-automatically by using a set of class labels as search keywords and collecting videos from web repositories (\eg, YouTube).
The result of retrieval is a collection of weakly annotated videos as each video is given its query keyword as video-level class label.
However, it is still not straightforward to learn semantic segmentation directly from weakly labeled videos due to ambiguous association between labels and frames.
The association is temporally ambiguous since only a subset of frames in a video is relevant to its class label.
Furthermore, although there are multiple regions exhibiting prominent motions, only a few among them might be relevant to the class label, which causes spatial ambiguity.
These ambiguities are ubiquitous in videos crawled automatically with no human intervention.

The key idea of this paper is to utilize both weakly annotated images and videos to learn a single DCNN for semantic segmentation.
Images are associated with clean class labels given manually, thus they can be used to alleviate the ambiguities in web-crawled videos. 
Also, it is easier to estimate shape and extent of object in videos thanks to motion cues available exclusively in them.
To exploit these complementary benefits of the two domains, we integrate techniques for discriminative object localization in images~\cite{Cam} and video segmentation~\cite{Inoutmap} into a single framework based on DCNN,
which generates reliable segmentation annotations from videos and learns semantic segmentation for image with the generated annotations.

The architecture of our DCNN is motivated by~\cite{transfernet} and consists of two parts, each of which has its own role: an encoder for image classification and discriminative localization~\cite{Cam}, and a decoder for image segmentation. 
The two parts of the network are trained separately with different data in our framework.
The encoder is first learned from a set of weakly annotated images.
It is in turn used to filter out irrelevant frames and identify discriminative regions in weakly annotated videos so that both temporal and spatial ambiguities of the videos are substantially reduced.
By incorporating the identified discriminative regions together with color and motion cues, spatio-temporal segments of object candidates are obtained from the videos by a well-established graph-based optimization technique. 
The video segmentation results are then used as segmentation annotations to train the decoder of our network.

The contributions of this paper are three-fold as follows.
\begin{itemize}
\item
We propose a weakly supervised semantic segmentation algorithm based on web-crawled videos.
Our algorithm exploits videos to simulate strong supervision missing in weakly annotated images, and utilizes images to eliminate noises in video retrieval and segmentation processes. 

\item
Our framework automatically collects video clips relevant to the target classes from web repositories so that it does not require human intervention to obtain extra supervision. 

\item 
We demonstrate the effectiveness of the proposed framework on the PASCAL VOC benchmark dataset, where it outperforms prior arts on weakly supervised semantic segmentation by a substantial margin.  
\end{itemize}

The rest of the paper is organized as follows.
We briefly review related work in Section~\ref{sec:relatedwork} and describe the details of the proposed framework in Section~\ref{sec:architecture}.
Section~\ref{sec:video_crawling} introduces data collection process.
Section~\ref{sec:experiments} illustrates experimental results on benchmark datasets.

\section{Related Work}
\label{sec:relatedwork}

Semantic segmentation has been rapidly improved in past few years, mainly due to emergence of powerful end-to-end learning framework based on DCNNs~\cite{Fcn,Deeplabcrf,Crfrnn,DeepParsing,Lin16,LRR,deconvnet}.
Built upon a fully-convolutional architecture~\cite{Fcn}, various approaches have been investigated to improve segmentation accuracy by integrating fully-connected CRF~\cite{Deeplabcrf,Crfrnn,DeepParsing,Lin16}, deep deconvolution network~\cite{deconvnet}, multi-scale processing~\cite{Deeplabcrf,LRR}, etc.
However, training a model based on DCNN requires pixel-wise annotations, which involves expensive and time-consuming procedures to obtain.
For this reasons, the task has been mainly investigated in small-scale datasets~\cite{Pascalvoc,Mscoco}. 

Approaches based on weakly supervised learning have been proposed to reduce annotation efforts in fully-supervised methods~\cite{Wsl,Wssl,Fcmil,Boxsup,Ccnn,sec,transfernet}. 
Among many possible choices, image-level labels are of the form requiring the minimum annotation cost thus have been widely used~\cite{Wsl, Fcmil,Ccnn,sec}.
Unfortunately, their results are far behind the fully-supervised methods due to missing supervision on segmentation. 
This gap is reduced by exploiting additional annotations such as point supervision~\cite{Bearman16}, scribble~\cite{scribblesup}, bounding box~\cite{Boxsup,Wssl}, masks from other class~\cite{transfernet}, 
but they lead to increased annotation cost that should be avoided in weakly supervised setting.
Instead of collecting extra cues from human annotator, we propose to retrieve and exploit web-videos, which offers motion cue useful for segmentation without the need of any human intervention in collecting such data. 
The idea of employing videos for semantic segmentation is new and has not been investigated properly except \cite{Tokmakov16}.
Our work is differentiated from \cite{Tokmakov16} by (i) exploiting complementary benefits in images and videos rather than directly learning from noisy videos, (ii) retrieving a large set of video clips from web repository rather than using a small number of manually collected videos. 
Our experimental results show that these differences lead to significant performance improvement.

Our work is closely related to webly-supervised learning~\cite{Chen15, levan, Neil, Tong15, Krause2016,TrackandTransfer,youtube-obj}, which aims to retrieve training examples from the resources on the Web.
The idea has been investigated in various tasks, such as concept recognition~\cite{Neil,levan,Chen15,Tong15}, object localization~\cite{levan,Neil,Tong15,TrackandTransfer}, and fine-grained categorization~\cite{Krause2016}.
The main challenge in this line of research is learning a model from noisy web data.
Various approaches have been employed such as curriculum learning~\cite{Chen15,Neil}, mining of visual relationship~\cite{levan}, semi-supervised learning with a small set of clean labels~\cite{Tong15}, etc. 
Our work addresses this issue using a model learned from another domain---we employ a model learned from a set of weakly annotated images to eliminate noises in web-crawled videos.

\begin{figure*}[!ht]
\centering
\includegraphics[width=0.932\linewidth] {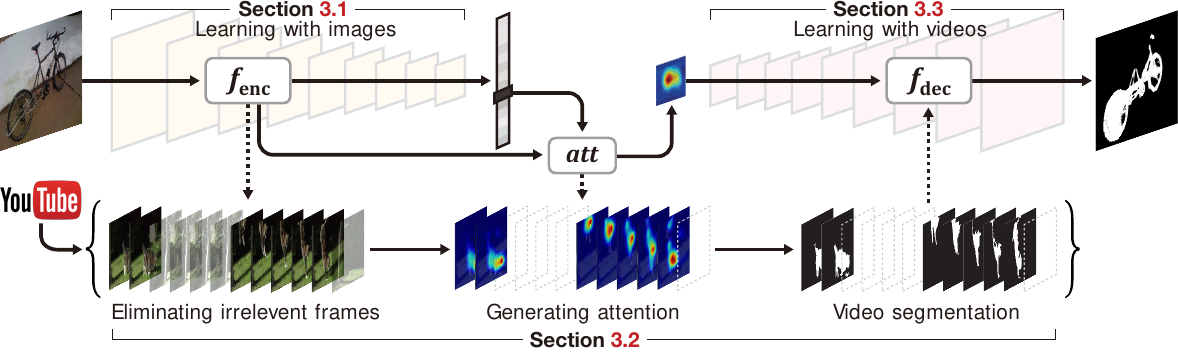}
\caption{
Overall framework of the proposed algorithm.
Our algorithm first learns a model for classification and localization from a set of weakly annotated images (Section~\ref{sec:attention_classification}). 
The learned model is used to eliminate noisy frames and generate coarse localization maps in web-crawled videos, where the per-pixel segmentation masks are obtained by solving a graph-based optimization problem (Section~\ref{sec:video_segmentation}).
The obtained segmentations are served as annotations to train a decoder (Section~\ref{sec:decoder}).
Semantic segmentation on still images is then performed by applying the entire network to images (Section~\ref{sec:inference}).
}
\label{fig:overview}
\end{figure*}

\section{Our Framework}
\label{sec:architecture}

The overall pipeline of the proposed framework is described in Figure~\ref{fig:overview}. 
We adopt a decoupled deep encoder-decoder architecture \cite{transfernet} as our model for semantic segmentation with a modification of its attention mechanism. 
In this architecture, the encoder $\fenc$ generates class prediction and a coarse attention map that identifies discriminative image regions for each predicted class, and the decoder $\fdec$ estimates a dense binary segmentation mask per class from the corresponding attention map.
We train each component of the architecture using different sets of data through the procedure below:
\begin{itemize}
\item 
Given a set of weakly annotated images, we train the encoder under a classification objective (Section~\ref{sec:attention_classification}).
\vspace{-0.1cm}
\item 
We apply the encoder to videos crawled on the Web to filter out frames irrelevant to their class labels, and generate a coarse attention map of the target class per remaining frame.
Spatio-temporal object segmentation is then conducted by solving an optimization problem incorporating the attention map with color and motion cues in each relevant interval of videos (Section~\ref{sec:video_segmentation}).
\vspace{-0.1cm}
\item 
We train the decoder by leveraging the segmentation labels obtained in the previous stage as supervision (Section~\ref{sec:decoder}).
\vspace{-0.1cm}
\item
Finally, semantic segmentation on still images is performed by applying the entire deep encoder-decoder network (Section~\ref{sec:inference}).
\end{itemize}
We also introduce a fully automatic method to retrieve relevant videos from web repositories (Section~\ref{sec:video_crawling}). 
This method enables us to construct a large collection of videos efficiently and effectively, which was critical to improved segmentation performance.
Following sections describe details of each step in our framework.

\subsection{Learning to Attend from Images}
\label{sec:attention_classification}

Let $\I$ be a dataset of weakly annotated images. 
An element of $\I$ is denoted by $(\x,\y)\in\I$, where $\x$ is an image and $\y\in\{0,1\}^C$ is a label vector for $C$ pre-defined classes.
We train the encoder $\fenc$ to recognize visual concepts under a classification objective by
\begin{equation}
\min_{\penc} \sum_{(\x,\y)\in\I} e_c(\y, \fenc(\x ; \penc)),
\label{eqn:obj_cls}
\end{equation}
where $\penc$ denotes parameters of $\fenc$, and $e_c$ is a cross-entropy loss for classification.  
For $\fenc$, we employ the pre-trained VGG-16 network~\cite{Vgg16} except its fully-connected layers, and place a new convolutional layer after the last convolutional layer of VGG-16 for better adaptation to our task. 
On the top of them, two additional layers, global average pooling followed by a fully-connected layer, are added to produce predictions on the class labels.
All newly added layers are randomly initialized.

Given the architecture and learned model parameters for $\fenc$, image regions relevant to each class are identified by Class Activation Mapping (CAM)~\cite{Cam}.
Let $F(\x)\in\R^{(w\cdot h)\times d}$ be output of the last convolutional layer of $\fenc$ given $\x$, and $W\in\R^{d\times C}$ the parameters for the fully-connected layer of $\fenc$, respectively, where $w,h$ and $d$ denote width, height and the number of channels of $F(\x)$.
Then for a class $c$, image regions relevant to the class are highlighted by CAM as follows:
%
\begin{equation}
\att^c = F(\x)\cdot\W\cdot\y^c,
\label{eqn:attention}
\end{equation}
where $\cdot$ is inner product and $\y^c\in \{0,1\}^C$ means a one-hot encoded vector for class $c$.
The output $\att^c\in\R^{w\cdot h}$ refers to an attention map for class $c$ and highlights local image regions relevant to class $c$.


\subsection{Generating Segmentation from Videos}
\label{sec:video_segmentation}

Our next step is to generate object segmentation masks from a set of weakly annotated videos using the encoder trained in the previous section.
Let $\V$ be a set of weakly annotated videos and $(\vvs,\y)\in\V$ an element in $\V$, where $\vvs=\{ \vv_1,...,\vv_T \}$ is a video composed of $T$ frames and $\y\in\{0,1\}^C$ is the label vector.
As in the image case, each video is associated with a label vector $\y$, but in this case it is a one-hot encoded vector since a single keyword is used to retrieve each video.

Having collected from the Web,
videos in $\V$ typically contain many frames irrelevant to associated labels.
Thus, segmenting objects directly from such videos may suffer from noises introduced by these frames. 
To address this issue, we measure class-relevance score of every frame $\vv$ in $\V$ with the learned encoder by $\y \cdot \fenc(\vv;\penc)$, 
and choose frames whose scores are larger than a threshold.
If more than 5 consecutive frames are chosen, we consider them as a single relevant video.
We construct a set of relevant videos ${\widehat \V}$, and perform object segmentation only on videos in ${\widehat \V}$.


The spatio-temporal segmentation of object is formulated by a graph-based optimization problem.
Let $\s_i^t$ be the $i$-th superpixel of frame $t$.
For each video $\vvs\in{\widehat \V}$, we construct a spatio-temporal graph $G=(\node,\edge)$, where a node corresponds to a superpixel $\s_i^t\in\node$, and the edges $\edge = \{\edge_s, \edge_t\}$ connect spatially adjacent superpixels $(\s_i^t, \s_j^t)\in\edge_s$ and temporally associated ones $(\s_i^t, \s_j^{t+1})\in\edge_t$.\footnote{We define a temporal edge between two superpixels from consecutive frames if they are connected by at least one optical flow~\cite{Bao2014}.}
Our goal is then reduced to estimating a binary label $l_i^t$ for each superpixel $\s_i^t$ in the graph $G$, where $l_i^t=1$ if $\s_i^t$ belongs to foreground (\ie, object) and $l_i^t=0$ otherwise.
The label estimation problem is formulated by the following energy minimization:
%
\begin{equation}
\min_\spl E(\spl) = E_u(\spl) + E_p(\spl),
\label{eqn:videoseg_obj}
\end{equation}
%
where $E_u$ and $E_p$ are unary and pairwise terms, respectively, and $\spl$ denotes labels of all superpixels in the video.
Details of the two energy terms are described below.

\begin{figure}[!t]
\centering
\includegraphics[width=.95\linewidth] {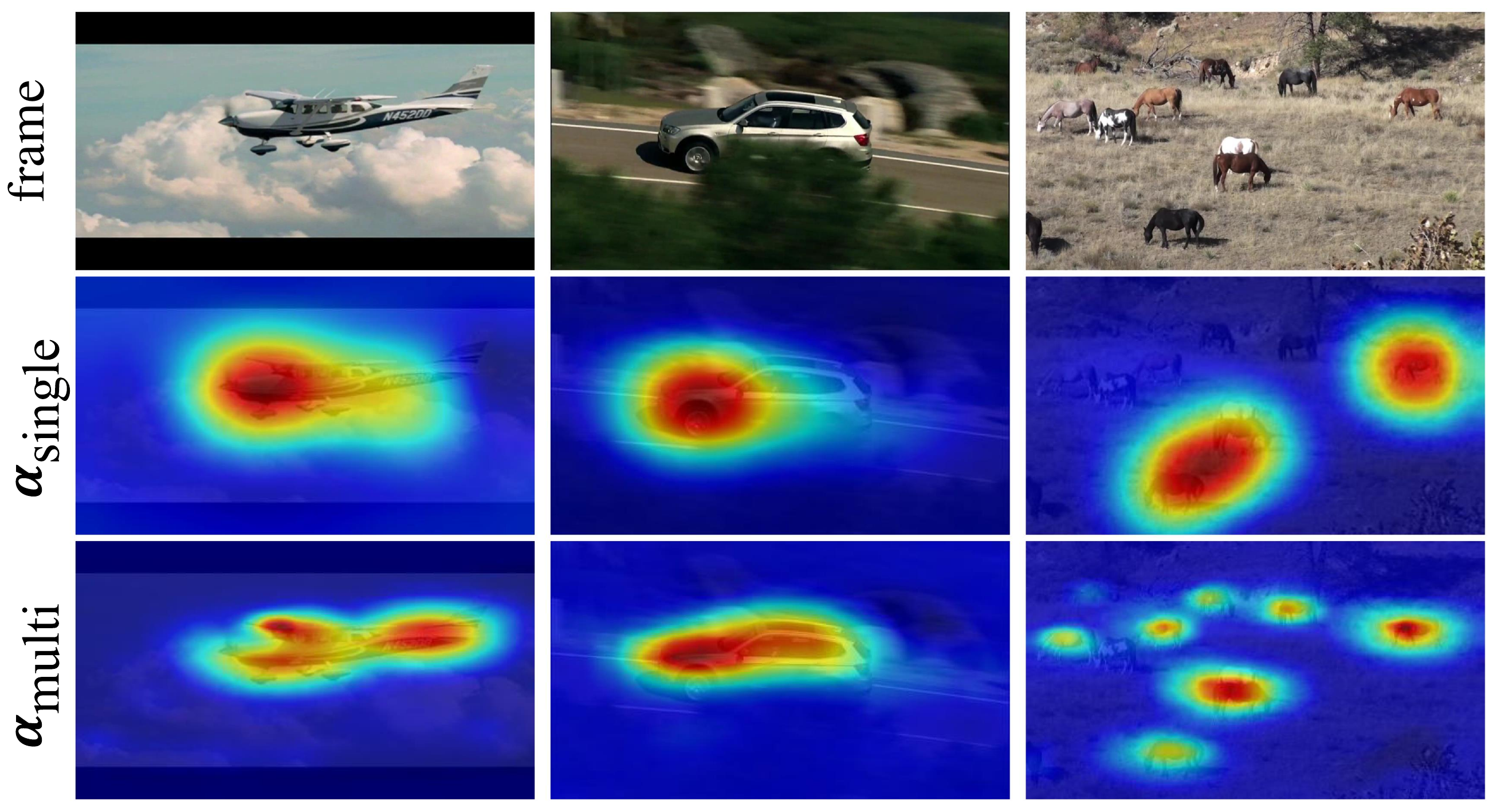}
\label{fig:cams}
\caption{
Qualitative examples of attention map on video frame.
(Top: video frame, Middle: attention with single scale input, Bottom: attention with multi-scale input.)
Although the encoder is trained on images, its attention maps effectively identify discriminative object parts in videos.
Also, multi-scale attention captures object parts and shapes better than its single scale counterpart.
}
\vspace{-0.2cm}
\end{figure}

\vspace{-0.13in}
\paragraph{Unary term.}
The unary term $E_u$ is a linear combination of three components that take various aspects of foreground object into account, and is given by
%
\begin{eqnarray}
E_u(\spl) &=& - \lambda_{\text{a}} \sum_{t,i} \log A^t_i (l^t_i) 
- \lambda_{\text{m}} \sum_{t,i} \log M^t_i (l^t_i) \nonumber \\
&& - \lambda_{\text{c}} \sum_{t,i} \log C^t_i (l^t_i),
\label{eqn:videoseg_unary}
\end{eqnarray}
where $A^t_i$, $C^t_i$ and $M^t_i$ denote the three components based on attention, appearance, and motion of superpixel $\s_i^t$, respectively. 
$\lambda_a$, $\lambda_c$, and $\lambda_m$ are weight parameters to control relative importance of the three terms.
We use the class-specific attention map obtained by Eq.~\eqref{eqn:attention} to compute the attention-based term $A^t_i$.
The attention map typically highlights discriminative parts of the object class, thus provides important evidences for video object segmentation.
To be more robust against scale variation of object, we compute multiple attention maps per frame by varying frame size.
After resizing them to the original frame size, we merge the maps through max-pooling over scale to obtain a single attention map per frame.
Figure~\ref{fig:cams} illustrates qualitative examples of such attention map.
$A^t_i$ is defined as attention over the superpixel $\s_i^t$, and calculated by aggregating the max-pooled attention values within the superpixel.

Although the attention term described above provide strong evidences for object localization, it tends to favor local discriminative parts of object since the model is trained under the classification objective in Eq.~\eqref{eqn:obj_cls}.
To better spread the localized attentions over the entire object area, we additionally take object appearance and motion into account.
The appearance term $C^t_i$ is implemented by a Gaussian Mixture Model (GMM).
Specifically, we estimate two GMMs based on RGB values of superpixels in the video, one for foreground and another for background. 
During GMM estimation, we first categorize superpixels into foreground and background by thresholding their attention values, and construct GMMs from the superpixels with their attention values as sample weights. 
The motion term $M^t_i$ returns higher value if the superpixel exhibiting more distinct motions is labeled as foreground.
We utilize inside-outside map from \cite{Inoutmap}, which identifies superpixels with distinct motion by estimating a closed curve following motion boundary. 
\vspace{-0.1cm}
\paragraph{Pairwise term.}
We employ the standard Potts model~\cite{Inoutmap,Rother04} to impose both spatial and temporal smoothness on inferred labels by
\begin{eqnarray}
E_p(\spl) &=& \hspace{-0.3cm} \sum_{(\s_i^t,\s_j^t)\in\edge_s}[l_i^t\neq l_j^t] \phi_s(\s_i^t,\s_j^t) \phi_c(\s_i^t,\s_j^t) +   \\
   && \hspace{-0.4cm} \sum_{(\s_i^t,\s_j^{t+1})\in\edge_t} [l_i^t\neq l_j^{t+1}] \phi_t(\s_i^t,\s_j^{t+1}) \phi_c(\s_i^t,\s_j^{t+1}) \nonumber
\label{eqn:videoseg_pairwise}
\end{eqnarray}
where $\phi_s$ and $\phi_c$ denote similarity metrics based on spatial location and color, respectively, and $\phi_t$ is the percentage of pixels connected by optical flows between the two superpixels.

\vspace{-0.13in}
\paragraph{Optimization.}
The Eq.~\eqref{eqn:videoseg_obj} is optimized efficiently by the Graph-cut algorithm.
The weight parameters are set to $\lambda_a=2$, $\lambda_m=1$, and $\lambda_c=2$.

\subsection{Learning to Segment from Videos}
\label{sec:decoder}
Given a set of generated segmentation annotations obtained in the previous section, we learn the decoder $\fdec$ for segmentation by
%
\begin{equation}
\min_{\pdec} \sum_{\vvs\in\widehat{\V}} \sum_{\vv\in\vvs} e_s(\spll_{\vv}^c, \fdec(\att^c_{\vv} ; \pdec)),
\label{eqn:decoder_obj}
\end{equation}
%
where $\pdec$ means parameters associated with the decoder, $\spll_{\vv}^c$ is a binary segmentation mask for class $c$ of frame $\vv$, 
and $e_s$ is a cross-entropy loss between prediction and the generated segmentation annotation.
Note that $\spll_{\vv}^c$ is computed from the segmentation labels $\spl$ estimated in the previous section.

We adopt the deconvolutional network \cite{deconvnet,decouplednet,transfernet} as our model for decoder $\fdec$, which is composed of multiple layers of deconvolution and unpooling. 
It takes the multi-scale attention map $\att^c_{\vv}$ of frame $\vv$ as an input, and produces a binary segmentation mask of class $c$ in the original resolution of the frame.
Since our multi-scale attention $\att^c_{\vv}$ already captures dense spatial configuration of object as illustrated in Figure~\ref{fig:cams}, our decoder does not require the additional densified-attention mechanism introduced in~\cite{transfernet}.
Note that the decoder is shared by all classes as no class label is involved in Eq.~\eqref{eqn:decoder_obj}.

The decoder architecture we adopt is well-suited to our problem for the following reasons.
First, the use of attention as input makes the optimization in Eq.~\eqref{eqn:decoder_obj} robust against incomplete segmentation annotations.
Because a video label identifies only one object class, segmentation annotations generated from the video ignore objects irrespective of the labeled class.
The decoder will get confused during training if such ignored objects are considered as background since they may be labeled as non-background in other videos.
By using the attention as input, the decoder does not care segmentation of such ignored objects and is thus trained more reliably.
Second, our decoder learns class-agnostic segmentation prior as it is shared by multiple classes during training~\cite{decouplednet}.
Since static objects (\eg, chair, table) are not well-separated from background by motion, their segmentation annotations are sometimes not plausible for training.
The segmentation prior learned from other classes is especially useful to improve the segmentation quality of such classes.

\subsection{Semantic Segmentation on Images}
\label{sec:inference}
Given encoder and decoder obtained by Eq. \eqref{eqn:obj_cls} and \eqref{eqn:decoder_obj}, semantic segmentation on still images is performed by the entire model. 
Specifically, given an input image $\x$, we first identify a set of class labels relevant to the image by thresholding the encoder output $\fenc(\x;\penc)$. 
Then for each identified label $c$, we compute attention map $\att^c$ by Eq.~\eqref{eqn:attention}, and generate corresponding foreground probability map from the output of decoder $\fdec( \att^c; \pdec )$.
The final per-pixel label is then obtained by taking pixel-wise maximum of $\fdec( \att^c; \pdec )$ for all identified classes.

\section{Video Retrieval from Web Repository}
\label{sec:video_crawling}
This section describes details of the video collection procedure.
Assume that we have a set of weakly annotated images $\I$, which is associated with predefined semantic classes.
Then for each class, we collect videos from YouTube using the class label as a search keyword to construct a set of weakly annotated videos $\V$.  
However, videos retrieved from YouTube are quite noisy in general because videos are often lacking side-information (\eg surrounding text) critical for text-based search, and class labels are usually too general to be used as search keywords (\eg person).
Although our algorithm is able to eliminate noisy frames and videos using the procedures described in Section~\ref{sec:video_segmentation}, examining all videos requires tremendous processing time and disk space, which should be avoided to construct a large-scale video data. 

We propose a simple, yet effective strategy that efficiently filters out noisy examples without looking at whole videos.
To this end, we utilize thumbnails and key-frames, which are global and local summaries of a video, respectively. 
In this strategy, we first download thumbnails rather than entire videos of search results, and compute classification scores of the thumbnails using the encoder learned from $\I$. 
Since a video is likely to contain informative frames if its thumbnail is relevant to the associated label, we download the video if classification score of its thumbnail is above a predefined threshold.
Then for each downloaded video, we extract key-frames\footnote{We utilize reference frames used to compress the video~\cite{mpeg} as key-frames for computational efficiency. This enables selection and extraction of informative video intervals without decompressing a whole video.} and compute their classification scores using the encoder to select only informative ones among them.
Finally, we extract frames within two seconds around each of selected key-frames to construct a video for $\V$. 
Videos in $\V$ may still contain irrelevant frames, which are handled by the procedure described in Section~\ref{sec:video_segmentation}.
We observe that videos collected by the above method are sufficiently clean and informative for learning.

\newcommand{\ROT}[1]{{\rotatebox[origin=c]{90}{#1}}}

\vspace{-0.1cm}
\section{Experiments}
\label{sec:experiments}

\vspace{-0.1cm}
\subsection{Implementation Details}
\label{sec:implementation_details}
\paragraph{Dataset.}
We employ the PASCAL VOC 2012 dataset~\cite{Pascalvoc} as the set of weakly annotated images $\I$, which contains 10,582 training images of 20 semantic categories.
The video retrieval process described in Section~\ref{sec:video_crawling} collects 4,606 videos and 960,517 frames for the raw video set $\V$ when we limit the maximum number of videos to 300 and select up to 15 key-frames per video.
The classification threshold for choosing relevant thumbnails and key-frames is set to 0.8, which favors precision more than recall.
\vspace{-0.13in}
\paragraph{Optimization.}
We implement the proposed algorithm based on Caffe~\cite{Caffe} library. 
We use Adam optimization~\cite{Adamsolver} to train our network with learning rate 0.001 and default hyper-parameter values proposed in \cite{Adamsolver}.
The size of mini-batch is set to 14.

\vspace{-0.1cm}
\subsection{Results on Semantic Segmentation}
\label{sec:results_semantic_segmentation}

%

This section presents semantic segmentation results on the PASCAL VOC 2012 benchmark~\cite{Pascalvoc}.
We employ \textit{comp6} evaluation protocol, and measure the performance based on mean Intersection Over Union (mIoU) between ground-truth and predicted segmentation.
\vspace{-0.13in}
\subsubsection{Internal Analysis}
\vspace{-0.1cm}
\label{sec:internal_analysis}
We first compare variants of our framework to verify impact of each component in the framework.
Table~\ref{tab:ablation_study} summarizes results of the internal analysis.

\vspace{-0.15in}
\paragraph{Impact of Separate Training.}
We compare our approach with \cite{Tokmakov16}, which also employs weakly annotated videos, but unlike ours, learns a whole model directly from the videos. 
For fair comparison, we train our model using the same set of videos from the YouTube-object dataset~\cite{youtube-obj}, which is collected manually from YouTube for 10 PASCAL object classes. 
Under the identical condition, our method substantially outperforms \cite{Tokmakov16} as shown in Table~\ref{tab:ablation_study}.
This result empirically demonstrates that our separate training strategy successfully takes advantage of the complementary benefits of image and video domains, while \cite{Tokmakov16} cannot.
%
\vspace{-0.15in}
\paragraph{Impact of Video Collection.}
Replacing a set of videos from \cite{youtube-obj} to the one collected from Section~\ref{sec:video_crawling} improves the performance by 6\% mIoU, although the videos are collected automatically with no human intervention.
It shows that (i) our model learns better object shapes from a larger amount of data and (ii) our video collection strategy is effective in retrieving informative videos from noisy web repositories.
\vspace{-0.15in}
\paragraph{Impact of Domain Adaptation.}
Examples in $\I$ and $\V$ have different characteristics: (i) They have different biases and data distributions, and 
(ii) images in $\I$ can be labeled by multiple classes while every video in $\V$ is annotated by a single class (\ie, search keyword). 
So we adapt our model trained on $\V$ to the domain of $\I$.
To this end, we apply the model to generate segmentation annotations of images in $\I$, and fine-tune the network using the generated annotations as strong supervision.
By the domain adaptation, the model learns context among multiple classes (\eg person rides bicycle) and different data distribution, which leads to the performance improvement by $3\%$ mIoU. 
%

%
\begin{table}[!t]
\centering
\caption{Comparisons between variants of the proposed framework on the PASCAL VOC 2012 validation set. 
DA stands for domain adaptation on still images. } \vspace{0.1cm}
\begin{tabular}{c|c|c|c}
method & video set & DA & mIoU \\
\hline
MCNN~\cite{Tokmakov16} & \cite{youtube-obj}& Y&38.1 \\
\hline
 & \cite{youtube-obj}& N& 49.2 \\
Ours & YouTube & N& 55.2 \\
 & YouTube & Y& {\bf 58.1} \\
\end{tabular}
\label{tab:ablation_study}
\end{table}

%
\vspace{-0.4cm}
\subsubsection{Comparisons to Other Methods}
\vspace{-0.2cm}
The performance of our framework is quantitatively compared with prior arts on weakly supervised semantic segmentation in Table~\ref{tab:voc_result_val} and~\ref{tab:voc_result_test}.
We categorize approaches based on types of annotations used in training.
\emph{Ours} denote our methods described in 4th row of Table~\ref{tab:ablation_study}.
Note that MCNN~\cite{Tokmakov16} utilizes manually collected videos~\cite{youtube-obj} where associations between labels and videos are not as ambiguous as those in our case.

Our method substantially outperforms existing approaches based on image-level labels, improving the state-of-the-art result by more than 7\% mIoU. 
Performance of our method is even as competitive as the approaches based on extra supervision, which rely on additional human intervention.
Especially, our method outperforms some approaches based on relatively stronger supervision (\eg, point supervision~\cite{Bearman16} and segmentation annotations of other classes~\cite{transfernet}).
These results show that segmentation annotations obtained from videos are sufficiently strong to simulate segmentation supervision missing in weakly annotated images.
Note that our method requires the same degree of human supervision with image-level labels since video retrieval is conducted fully automatically in the proposed framework.

\begin{figure*}[!ht]
\small
 \hspace{1cm}Input Image\hspace{1.7cm} Ground-truth \hspace{1.9cm} SEC~\cite{sec} \hspace{1.9cm} MCNN~\cite{Tokmakov16}\hspace{2.3cm} Ours
\vspace{-0.18cm}
\begin{center}
\includegraphics[width=0.195\linewidth] {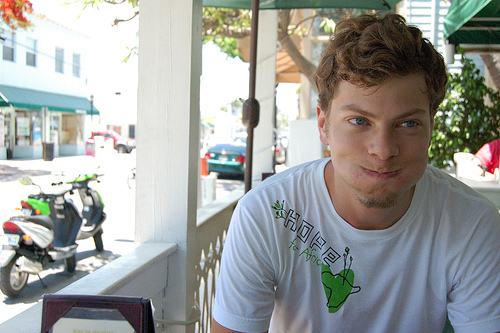}
\includegraphics[width=0.195\linewidth] {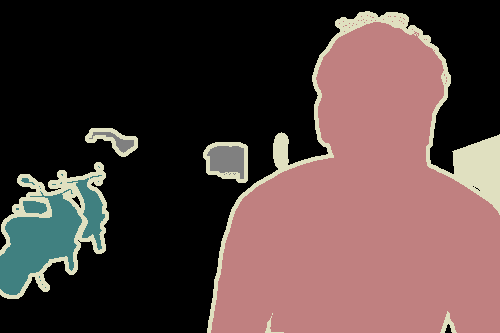}
\includegraphics[width=0.195\linewidth] {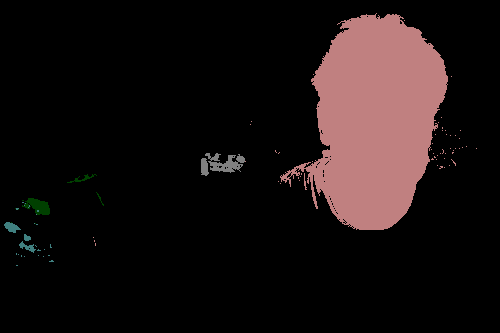}
\includegraphics[width=0.195\linewidth] {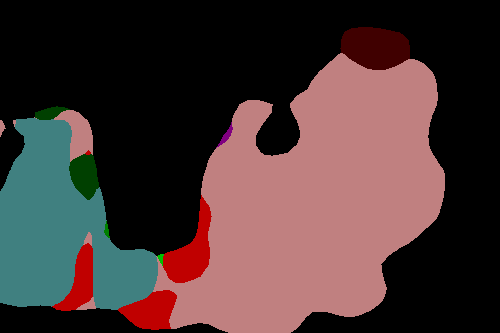}
\includegraphics[width=0.195\linewidth] {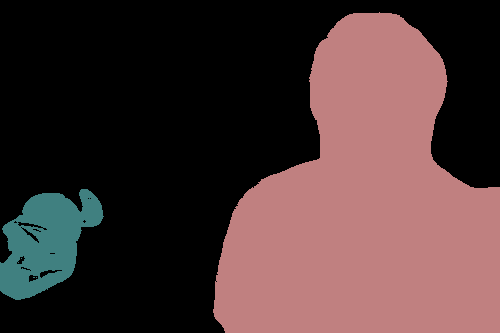}\\
\includegraphics[width=0.195\linewidth] {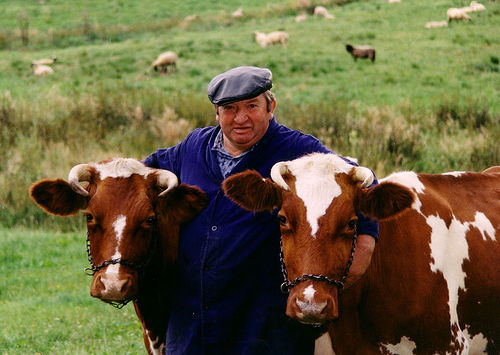}
\includegraphics[width=0.195\linewidth] {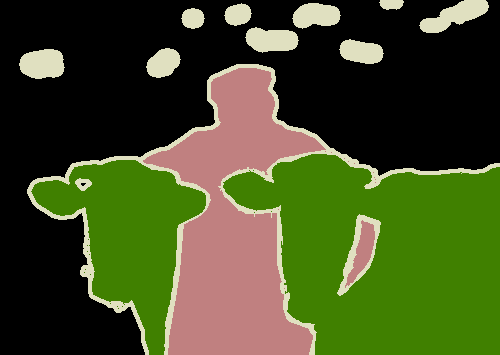}
\includegraphics[width=0.195\linewidth] {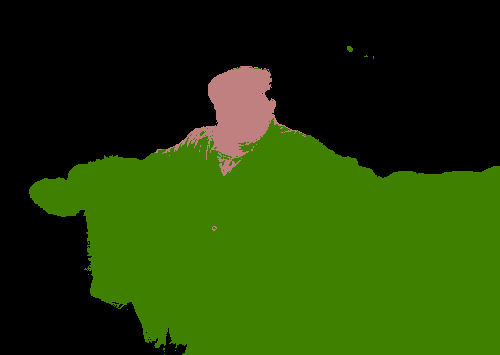}
\includegraphics[width=0.195\linewidth] {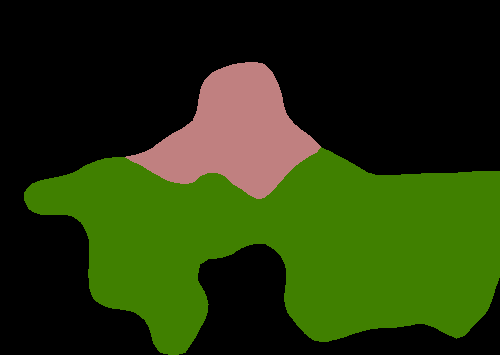}
\includegraphics[width=0.195\linewidth] {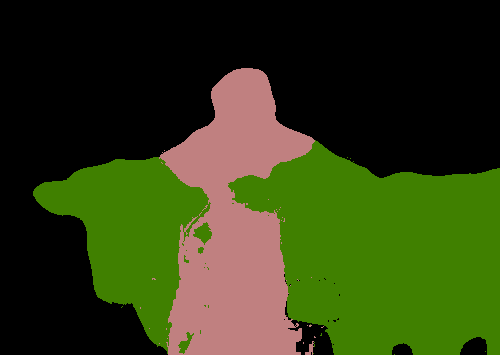}\\
\includegraphics[width=0.195\linewidth] {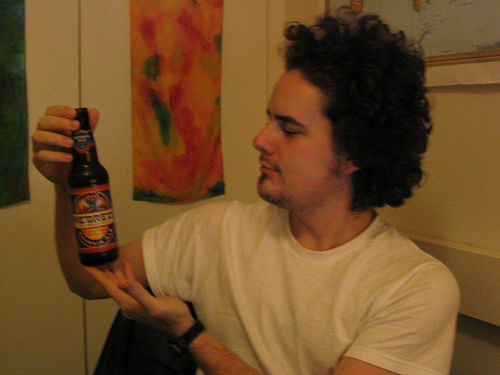}
\includegraphics[width=0.195\linewidth] {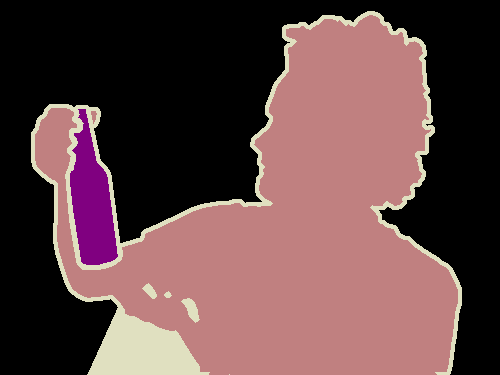}
\includegraphics[width=0.195\linewidth] {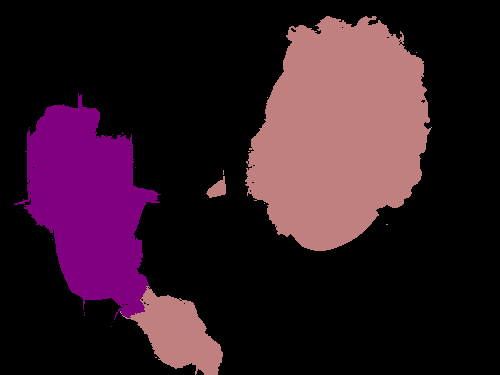}
\includegraphics[width=0.195\linewidth] {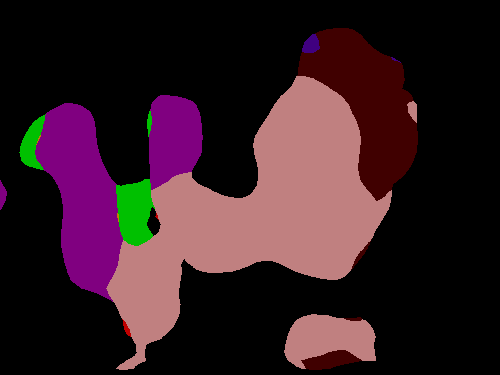}
\includegraphics[width=0.195\linewidth] {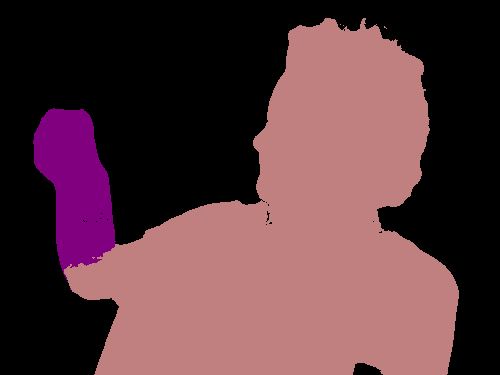}\\
\includegraphics[width=0.195\linewidth] {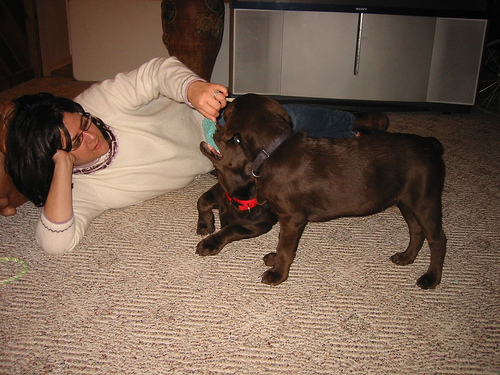}
\includegraphics[width=0.195\linewidth] {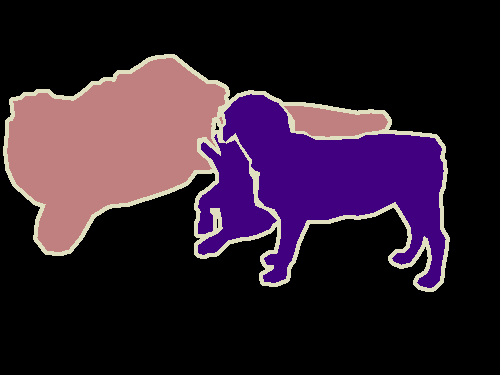}
\includegraphics[width=0.195\linewidth] {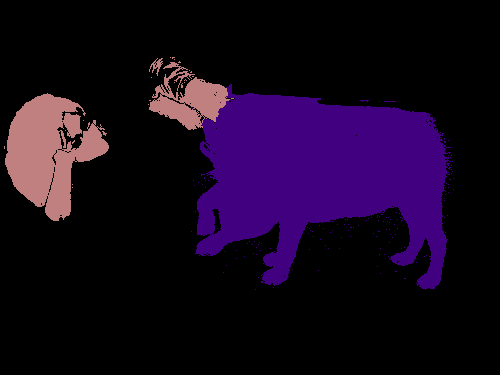}
\includegraphics[width=0.195\linewidth] {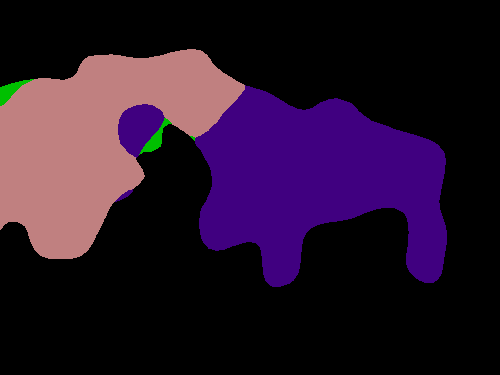}
\includegraphics[width=0.195\linewidth] {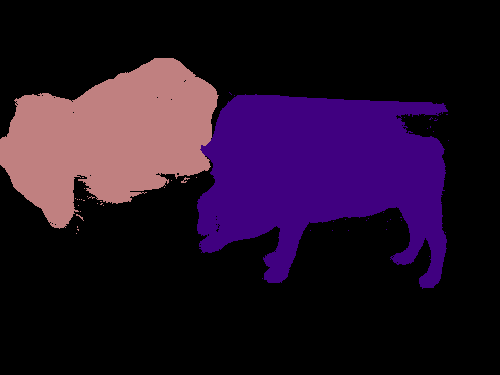}\\
\includegraphics[width=0.195\linewidth] {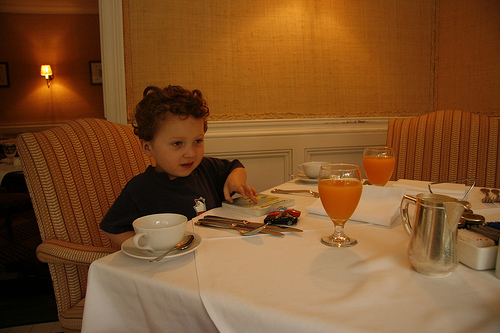}
\includegraphics[width=0.195\linewidth] {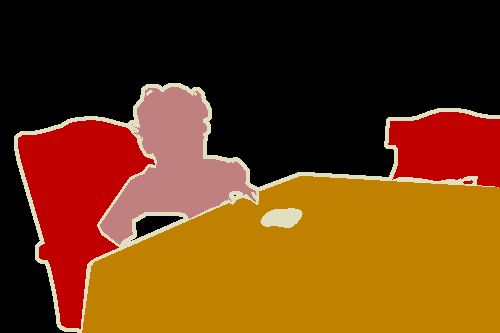}
\includegraphics[width=0.195\linewidth] {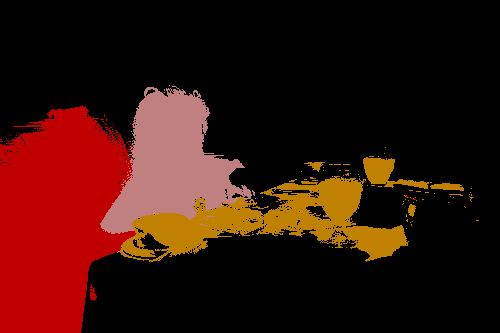}
\includegraphics[width=0.195\linewidth] {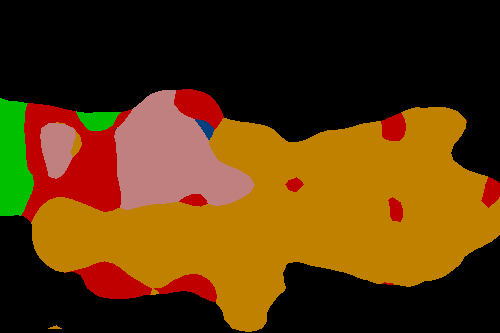}
\includegraphics[width=0.195\linewidth] {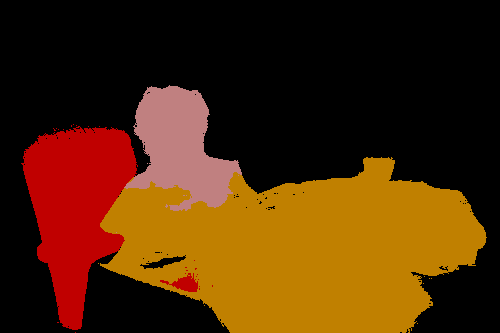}\\
\includegraphics[width=0.195\linewidth] {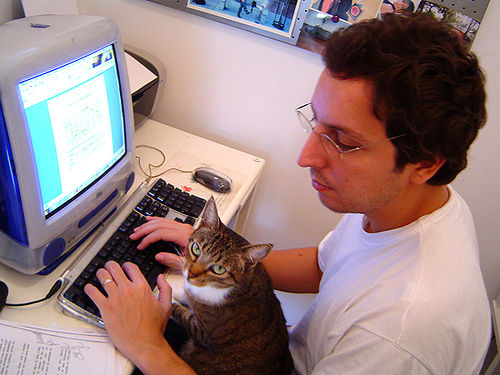}
\includegraphics[width=0.195\linewidth] {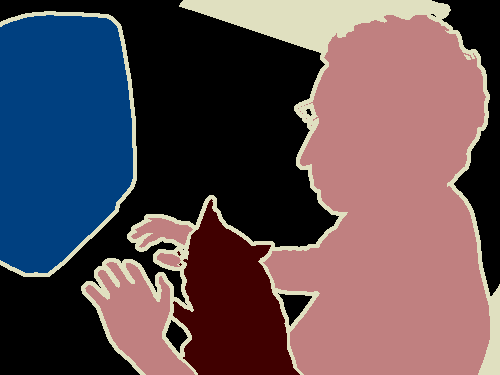}
\includegraphics[width=0.195\linewidth] {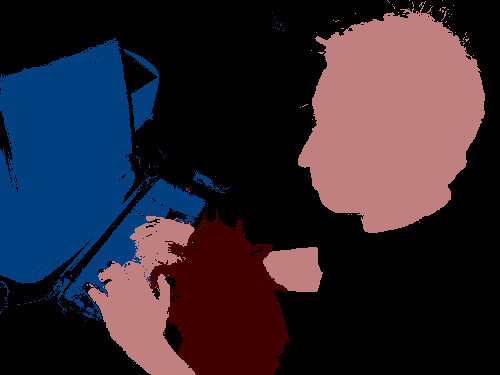}
\includegraphics[width=0.195\linewidth] {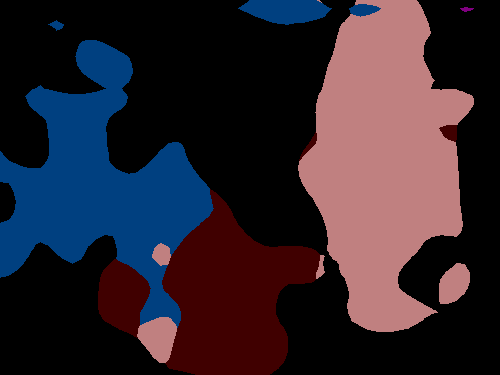}
\includegraphics[width=0.195\linewidth] {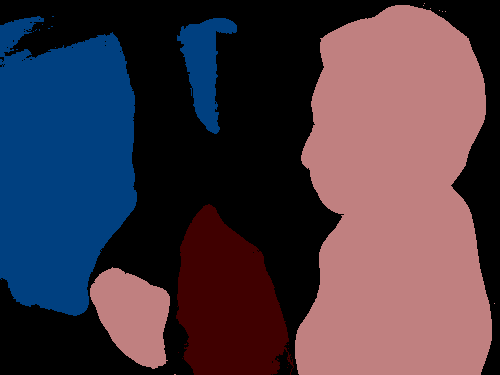}\\
\end{center}
\caption{
Qualitative results on the PASCAL VOC 2012 validation images. 
SEC~\cite{sec} is the state of the art among the approaches relying only on image-level class labels, and MCNN~\cite{Tokmakov16} exploits videos as an additional source of training data as ours does. 
Compared to these approaches, our method captures object boundary more accurately and covers larger object area. 
} 
\label{fig:qualitative_voc}
\end{figure*}

Figure~\ref{fig:qualitative_voc} illustrates qualitative results.
Compared to approaches based only on image labels, our method tends to produce more accurate predictions on object location and boundary.

%
\begin{table*}[!t] \footnotesize
\centering
\caption{Evaluation results on the PASCAL VOC 2012 \textit{validation} set.} \vspace{0.1cm}
\begin{tabular}
{
@{}C{2.7cm}@{}|@{}C{0.68cm}@{}C{0.66cm}@{}C{0.66cm}@{}C{0.66cm}@{}C{0.66cm}@{}C{0.66cm}@{}C{0.66cm}@{}C{0.66cm}@{}C{0.66cm}@{}C{0.66cm}@{}C{0.66cm}@{}C{0.66cm}@{}C{0.66cm}@{}C{0.66cm}@{}C{0.73cm}@{}C{0.75cm}@{}C{0.66cm}@{}C{0.66cm}@{}C{0.66cm}@{}C{0.66cm}@{}C{0.66cm}@{}|@{}C{0.75cm}@{}
}
\hline
Method&bkg&aero&bike&bird&boat&bottle&bus&car&cat&chair&cow&table&dog&horse&mbk&person&plant&sheep&sofa&train&tv&mean\\
\hline
\textbf{Image labels:} \raggedright & & & & & & & & & & & & & & & & & & & & & & \\
EM-Adapt~\cite{Wssl} \raggedright & 67.2 & 29.2 & 17.6 & 28.6 & 22.2 & 29.6 & 47.0 & 44.0 & 44.2 & 14.6 & 35.1 & 24.9 & 41.0 & 34.8 & 41.6 & 32.1 & 24.8 & 37.4 & 24.0 & 38.1 & 31.6 & 33.8 \\ 
CCNN~\cite{Ccnn} \raggedright & 68.5 & 25.5 & 18.0 & 25.4 & 20.2 & 36.3 & 46.8 & 47.1 & 48.0 & 15.8 & 37.9 & 21.0 & 44.5 & 34.5 & 46.2 & 40.7 & 30.4 & 36.3 & 22.2 & 38.8 & 36.9 & 35.3 \\ 
MIL+seg~\cite{Wsl} \raggedright & 79.6 & 50.2 & 21.6 & 40.9 & 34.9 & 40.5 & 45.9 & 51.5 & 60.6 & 12.6 & 51.2 & 11.6 & 56.8 & 52.9 & 44.8 & 42.7 & 31.2 & 55.4 & 21.5 & 38.8 & 36.9 & 42.0 \\ 
SEC~\cite{sec} \raggedright & 82.4& 62.9& 26.4& 61.6& 27.6& 38.1& 66.6& 62.7& 75.2& 22.1& 53.5& 28.3& 65.8& 57.8& 62.3& 52.5& 32.5& 62.6& 32.1& 45.4& 45.3& 50.7 \\
\hline
\textbf{+Extra annotations:} \raggedright & & & & & & & & & & & & & & & & & & & & & & \\
Point supervision~\cite{Bearman16} \raggedright & 80.0& 49.0& 23.0& 39.0& 41.0& 46.0& 60.0& 61.0& 56.0& 18.0& 38.0& 41.0& 54.0& 42.0& 55.0& 57.0& 32.0& 51.0& 26.0& 55.0& 45.0& 46.0\\
Bounding box~\cite{Wssl} \raggedright & -& -& -& -& -& -& -& -& -& -& -& -& -& -& -& -& -& -& -& -& -& 58.5\\
Bounding box~\cite{Boxsup} \raggedright & -& -& -& -& -& -& -& -& -& -& -& -& -& -& -& -& -& -& -& -& -& 62.0\\
Scribble~\cite{scribblesup} \raggedright & -& -& -& -& -& -& -& -& -& -& -& -& -& -& -& -& -& -& -& -& -& 63.1\\
Transfer learning~\cite{transfernet} \raggedright &85.3 &68.5 &26.4 &69.8 &36.7 &49.1 &68.4 &55.8 &77.3 &6.2 &75.2 &14.3 &69.8 &71.5 &61.1 &31.9 &25.5 &74.6 &33.8 &49.6 &43.7 & 52.1\\
\hline
\textbf{+Videos} (unannotated): \raggedright & & & & & & & & & & & & & & & & & & & & & & \\
MCNN~\cite{Tokmakov16} \raggedright &77.5 &47.9 &17.2 &39.4 &28.0 &25.6 &52.7 &47.0 &57.8 &10.4 &38.0 &24.3 &49.9 &40.8 &48.2 &42.0 &21.6 &35.2 &19.6 &52.5 &24.7 &38.1 \\
Ours \raggedright &87.0 &69.3 &32.2 &70.2 &31.2 &58.4 &73.6 &68.5 &76.5 &26.8 &63.8 &29.1 &73.5 &69.5 &66.5 &70.4 &46.8 &72.1 &27.3 &57.4 &50.2 &{\bf 58.1} \\
\hline
\end{tabular}
\label{tab:voc_result_val}
\end{table*}
\begin{table*}[!t] \footnotesize
\centering
\caption{Evaluation results on the PASCAL VOC 2012 \textit{test} set.} \vspace{0.1cm}
\begin{tabular}
{
@{}C{2.7cm}@{}|@{}C{0.68cm}@{}C{0.66cm}@{}C{0.66cm}@{}C{0.66cm}@{}C{0.66cm}@{}C{0.66cm}@{}C{0.66cm}@{}C{0.66cm}@{}C{0.66cm}@{}C{0.66cm}@{}C{0.66cm}@{}C{0.66cm}@{}C{0.66cm}@{}C{0.66cm}@{}C{0.73cm}@{}C{0.75cm}@{}C{0.66cm}@{}C{0.66cm}@{}C{0.66cm}@{}C{0.66cm}@{}C{0.66cm}@{}|@{}C{0.75cm}@{}
}
\hline
Method&bkg&aero&bike&bird&boat&bottle&bus&car&cat&chair&cow&table&dog&horse&mbk&person&plant&sheep&sofa&train&tv&mean\\
\hline
\textbf{Image labels:} \raggedright & & & & & & & & & & & & & & & & & & & & & \\
EM-Adapt~\cite{Wssl} \raggedright & 76.3 & 37.1 & 21.9 & 41.6 & 26.1 & 38.5 & 50.8 & 44.9 & 48.9 & 16.7 & 40.8 & 29.4 & 47.1 & 45.8 & 54.8 & 28.2 & 30.0 & 44.0 & 29.2 & 34.3 & 46.0 & 39.6 \\
CCNN~\cite{Ccnn} \raggedright & 70.1 & 24.2 & 19.9 & 26.3 & 18.6 & 38.1 & 51.7 & 42.9 & 48.2 & 15.6 & 37.2 & 18.3 & 43.0 & 38.2 & 52.2 & 40.0 & 33.8 & 36.0 & 21.6 & 33.4 & 38.3 & 35.6 \\
MIL+seg~\cite{Wsl} \raggedright & 78.7 & 48.0 & 21.2 & 31.1 & 28.4 & 35.1 & 51.4 & 55.5 & 52.8 & 7.8 & 56.2 & 19.9 & 53.8 & 50.3 & 40.0 & 38.6 & 27.8 & 51.8 & 24.7 & 33.3 & 46.3 & 40.6 \\
SEC~\cite{sec} \raggedright & 83.5& 56.4& 28.5& 64.1& 23.6& 46.5& 70.6& 58.5& 71.3& 23.2& 54.0& 28.0& 68.1& 62.1& 70.0& 55.0& 38.4& 58.0& 39.9& 38.4& 48.3& 51.7\\
\hline
\textbf{+Extra annotations:} \raggedright & & & & & & & & & & & & & & & & & & & & & & \\
Point supervision~\cite{Bearman16} \raggedright & 80.0& 49.0& 23.0& 39.0& 41.0& 46.0& 60.0& 61.0& 56.0& 18.0& 38.0& 41.0& 54.0& 42.0& 55.0& 57.0& 32.0& 51.0& 26.0& 55.0& 45.0& 46.0\\
Bounding box~\cite{Wssl} \raggedright & -& -& -& -& -& -& -& -& -& -& -& -& -& -& -& -& -& -& -& -& -& 60.4\\
Bounding box~\cite{Boxsup} \raggedright & -& -& -& -& -& -& -& -& -& -& -& -& -& -& -& -& -& -& -& -& -& 64.6\\
Transfer learning~\cite{transfernet} \raggedright &85.7 &70.1 &27.8 &73.7 &37.3 &44.8 &71.4 &53.8 &73.0 &6.7 &62.9 &12.4 &68.4 &73.7 &65.9 &27.9 &23.5 &72.3 &38.9 &45.9 &39.2&51.2\\
\hline
\textbf{+Videos} (unannotated): \raggedright & & & & & & & & & & & & & & & & & & & & & \\
MCNN~\cite{Tokmakov16} \raggedright & 78.9 &48.1 &17.9 &37.9 &25.4 &27.5 &53.4 &48.8 &58.3 &9.9 &43.2 &26.6 &54.9 &49.0 &51.1 &42.5 &22.9 &39.3 &24.2 &50.2 &25.9 &39.8 \\
Ours \raggedright &87.2 &63.9 &32.8 &72.4 &26.7 &64.0 &72.1 &70.5 &77.8 &23.9 &63.6 &32.1 &77.2 &75.3 &76.2 &71.5 &45.0 &68.8 &35.5 &46.2 &49.3 &{\bf 58.7} \\
\hline
\end{tabular}
\label{tab:voc_result_test}
\end{table*}

\subsection{Results on Video Segmentation}
\label{sec:results_video_segmentation}
%
\begin{table} 
\centering
\caption{Evaluation results of video segmentation performance on the YouTube-object benchmark.} \vspace{0.1cm}
\begin{tabular}{c|c|c|c}
method & extra data&class avg.& video avg.\\
\hline
\cite{Tang13}& -&23.9& 22.8\\
\cite{Inoutmap}& -&46.8& 43.2\\
\hline
\cite{Zhang15}& bounding box&54.1& 52.6\\
\cite{Drayer16}& bounding box&56.2& 55.8\\
\hline
Ours & image label&{\bf 58.6}& {\bf 57.1}
\vspace{-0.4cm}
\end{tabular}
\label{tab:youtube_obj_results}
\end{table}
%

To evaluate the quality of video segmentation results obtained by the proposed framework, we compare our method with state-of-the-art video segmentation algorithms on the YouTube-object benchmark dataset~\cite{youtube-obj}.
We employed segmentation ground-truths from \cite{Jain14} for evaluation, which provides a binary segmentation masks at every 10 frames for selected video intervals.
Following protocols in the previous work, we measure the performance based on mIoU over categories and videos.

The summary results are shown in Table~\ref{tab:youtube_obj_results}. 
Our method substantially outperforms previous approaches based only on low-level cues such as motion and appearance, since the attention map we employ provides robust and semantically meaningful estimation of object location in video.
Interestingly, our method outperforms approaches using object detector trained on bounding box annotations~\cite{Zhang15, Drayer16} that require stronger supervision than image-level labels. 
This may be because attention map produced by our method provides more fine-grained localization of an object than coarse bounding box predicted by object detector.

\begin{figure*}[!ht]
\centering
\includegraphics[width=0.196\linewidth] {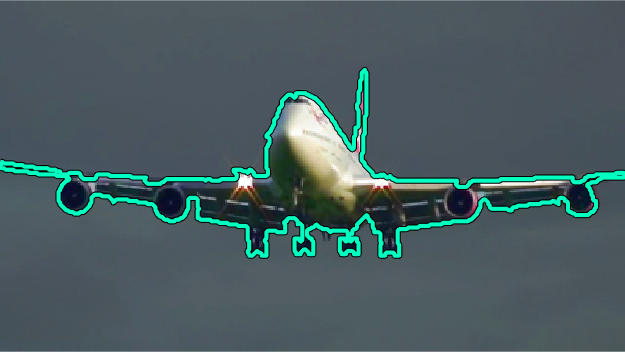}
\includegraphics[width=0.195\linewidth] {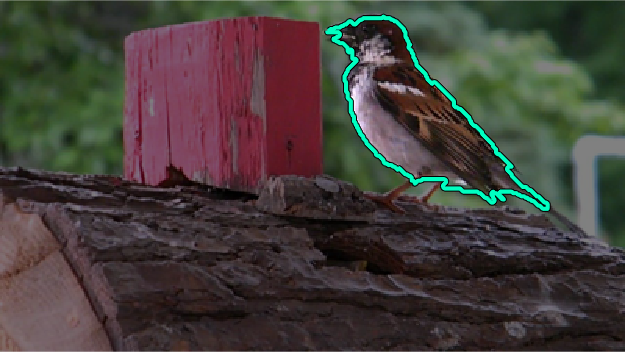}
\includegraphics[width=0.195\linewidth] {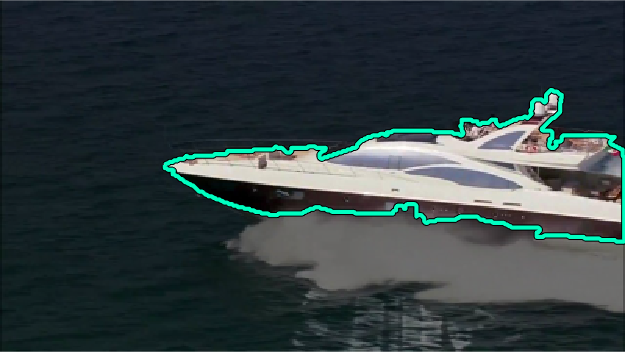}
\includegraphics[width=0.195\linewidth] {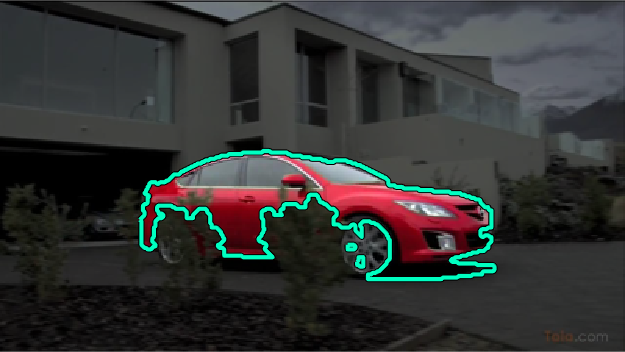}
\includegraphics[width=0.195\linewidth] {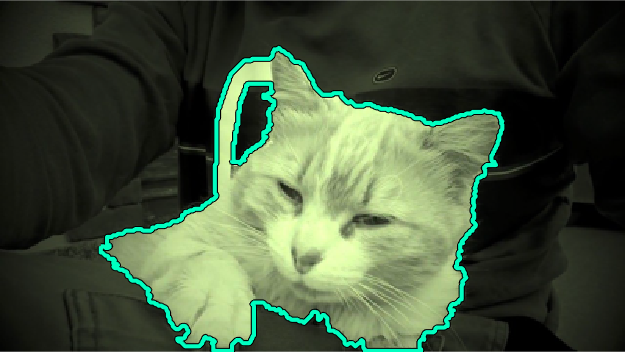}\\
\includegraphics[width=0.195\linewidth] {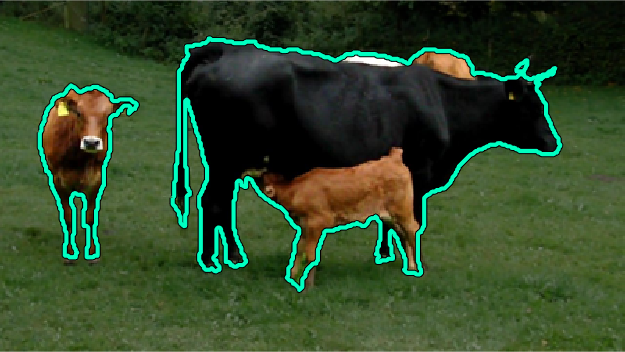}
\includegraphics[width=0.195\linewidth] {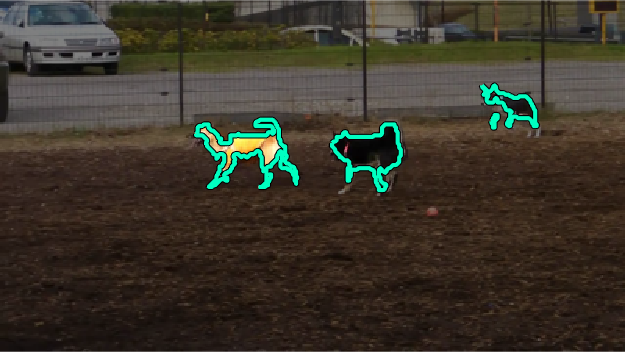}
\includegraphics[width=0.195\linewidth] {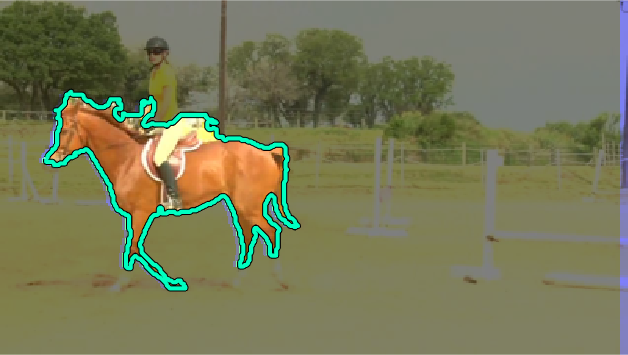}
\includegraphics[width=0.195\linewidth] {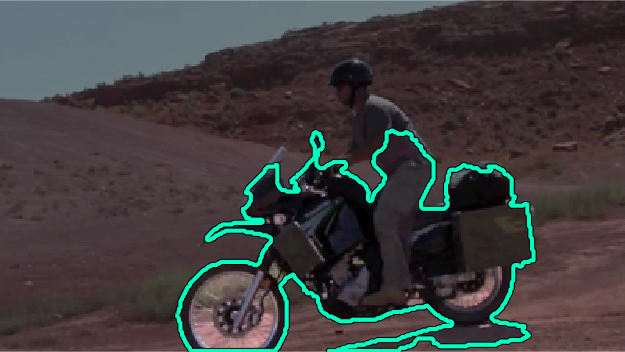}
\includegraphics[width=0.195\linewidth] {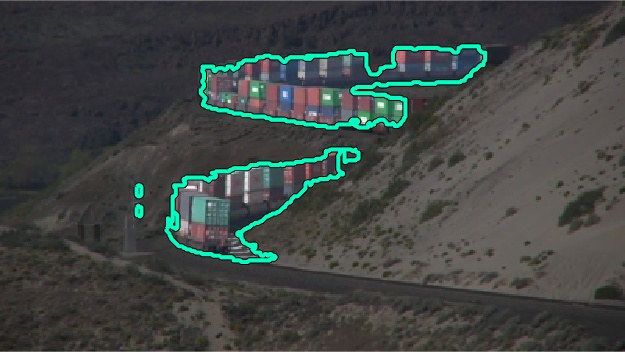}
\caption{Qualitative results of the proposed method on the YouTube-object dataset.
Our method segments objects successfully in spite of challenges like occlusion (\eg, \emph{car}, \emph{train}), background clutter (\eg, \emph{bird}, \emph{car}), multiple instances (\eg, \emph{cow}, \emph{dog}), and irrelevant objects that cannot be distinguished from target object by motion (\eg \emph{people} riding \emph{horse} and \emph{motorbike}).
}
\label{fig:qualitative_youtube}
\end{figure*}

Figure~\ref{fig:qualitative_youtube} illustrates qualitative results of the proposed approach.
Our method generates accurate segmentation masks under various challenges in videos, 
such as occlusion, background clutter, objects of other classes, and so on.
More comprehensive qualitative results are available at our project webpage\footnote{\url{http://cvlab.postech.ac.kr/research/weaksup_video/}}.

\vspace{-0.1cm}
\section{Conclusion}
\label{sec:conclusion}

We propose a novel framework for weakly supervised semantic segmentation based on image-level class labels only.
The proposed framework retrieves relevant videos automatically from the Web, and generates fairly accurate object masks of the classes from the videos to simulate supervision for semantic segmentation.
For reliable object segmentation in video, our framework first learns an encoder from weakly annotated images to predict attention map, and incorporates the attention with motion cues in videos to capture object shape and extent more accurately.
The obtained masks are then served as segmentation annotations to learn a decoder for segmentation.
Our method outperformed previous approaches based on the same level of supervision, and as competitive as the approaches relying on extra supervision.

\vspace{-0.2cm}
\paragraph{Acknowledgments}
\small
This work was partly supported by IITP grant (2014-0-00147 and 2016-0-00563), NRF grant (NRF-2011-0031648), DGIST Faculty Start-up Fund (2016080008), NSF CAREER IIS-1453651, ONR N00014-13-1-0762, and a Sloan Research Fellowship.

{\small
\bibliographystyle{ieee}
\bibliography{egbib}
}

\end{document}